\RequirePackage{fix-cm}
\documentclass[smallextended]{svjour3}       
\smartqed  
\usepackage{graphicx}
\usepackage{float} 
\usepackage{natbib} 
\usepackage{comment}
\usepackage{xcolor}
\usepackage{amsmath}
\usepackage[misc, geometry]{ifsym}
\usepackage{url}

\newcommand*{\affaddr}[1]{#1} 
\newcommand*{\affmark}[1][*]{\textsuperscript{#1}}
\begin{document}

\title{Using attention methods to predict judicial outcomes
}


\author{Vithor Gomes Ferreira Bertalan   \protect\affmark[1]       \and
        Evandro Eduardo Seron Ruiz  \affmark[2]
}

\authorrunning{Vithor Bertalan \and Evandro E.~S. Ruiz} 

\institute{  Vithor Gomes Ferreira Bertalan \at
              \email{vithor.bertalan@polymtl.ca}           
           \and
           Evandro Eduardo Seron Ruiz \at
              \email{evandro@usp.br}\\ \\
              \affaddr{\affmark[1]Département de Génie Informatique et Génie Logiciel de l'École Polytechnique de Montréal, Université de Montréal, Montréal, Canada.}\\
\affaddr{\affmark[2]Departamento de Computação e Matemática da FFCLRP, Universidade de São Paulo, Ribeirão Preto, Brazil.}\\
}

\date{Received: date / Accepted: date}

\maketitle

\begin{abstract}
The prediction of legal judgments is one of the most recognized fields in Natural Language Processing (NLP), Artificial Intelligence (AI), and Law combined. By legal prediction, we mean intelligent systems capable of predicting specific judicial characteristics such as the judicial outcome, the judicial class, and the prediction of a particular case. In this study, we used an artificial intelligence classifier to predict the decisions of Brazilian courts. To this end, we developed a text crawler to extract data from official Brazilian electronic legal systems, consisting of two datasets of cases of second-degree murder and active corruption. We applied various classifiers, such as Support Vector Machines (SVM), Neural Networks, and others, to predict judicial outcomes by analyzing text features from the dataset. Our research demonstrated that Regression Trees, Gated Recurring Units, and Hierarchical Attention Networks tended to have higher metrics across our datasets. As the final goal, we searched the weights of one of the algorithms, Hierarchical Attention Networks, to find samples of the words that might be used to acquit or convict defendants based on their relevance to the algorithm. 
\keywords{Legal Prediction \and Hierarchial Attention Networks \and Natural Language Processing \and Machine Learning}
\end{abstract}

\section{Introduction}
\label{intro}
Computer science has redefined many spheres of experience in our time. Many computer science subfields, such as Natural Language Processing (NLP), have steadily improved many professional and scientific activities. NLP is a range of computational techniques based on theoretical principles for automatic human language analysis and representation \citep{Cambria2014}. One of the areas of human knowledge that most depend on text is the law. Many legal articles, contracts, deeds, orders/judgments/decrees, statutes, court decisions, and appeals are written daily, and many different positions, such as lawyers, judges, and defendants, have different needs that intelligent systems can provide.

Thus, it is reasonable to consider that Artificial Intelligence (AI) can also be used to optimize the daily tasks of legal professionals. Although legal methods are long established, the field of law AI is still developing. Moreover, when analyzing the state-of-the-art, we saw that much research covers only the English language. Analyzes in Brazilian Portuguese are still in their infancy and pose an exciting challenge for new applications~\citep{oliveira2020indicators}. 

This study's main objective was to develop a framework for predicting judicial outcomes in the São Paulo Justice Court\footnote{Tribunal de Justiça do Estado de São Paulo, Brasil, TJSP.}. Taking into account the number of legal processes, the São Paulo Justice Court is the most extensive court law in the world \footnote{https://www.tjsp.jus.br/QuemSomos (in Brazilian Portuguese)}. A computational forecasting model that provides a satisfactory result for this large judicial court could be beneficial and, perhaps after fine-tuning, could also be applied to other courts. Recent years have focused on predicting the outcome of courts using NLP and AI methods. Researchers have dedicated their research to predicting the outcome of judicial cases \citep{sun2020legal, antos2021practical}. However, as far as we know, there are no studies in Brazilian Portuguese or Brazilian courts involving this intention as of 2022.

We initially built a text crawler to collect data from legal outcomes. This data set constituted an annotated corpus for training and evaluating the proposed prediction system. A pre-processing phase was necessary to extract the features from the data set. The selected features have been incorporated into machine learning frameworks such as neural networks and support vector machines to preview the court's outcomes.

The rest of this work is structured as follows. In Section~\ref{backgd}, we will present related work within the theoretical background and indicate previous research and knowledge. In Section~\ref{meth}, we explain the methodology used in our research and how it was implemented. Section~\ref{exp} describes the experiments conducted on the labeled dataset. Finally, in Section~\ref{conc}, we present our conclusions and future work possibilities.

\section{Theoretical Background}
\label{backgd}
This section describes an overview of the current work on artificial intelligence and natural language processing applied to legal texts. We also present some of the inherent challenges associated with this field.

\subsection{Natural Language Processing, Artificial Intelligence, and Law}

According to \cite{info:doi/10.2196/16816}, Natural Language Processing (NLP) is the ability of machines to understand and explain how humans write and talk. However, to understand NLP and AI and how they interact with the law, we must first understand the last. The law, as defined by \cite{Le2015}, is the system of rules that guarantees peace, personal freedom, and social justice by regulating human behavior. Legal documents are, according to the authors, documents that establish a contractual relationship or give rights. Since NLP works well with documents, applying NLP in law may seem like a perfect match. 

However, regardless of all recent technological advances, \cite{Surden2014} argues that modern AI algorithms cannot replicate most human intellectual abilities and fall far short of advanced cognitive processes, such as logical reasoning, which are fundamental to legal practice. According to \cite{Branting2017a}, the main reason is that scaling artificial systems to comprehend the characteristics of complex and dynamic real-world legal systems is a difficult task. The authors have proposed two main challenges: the problem of effectively and verifiably representing legal texts as local expressions and the difficulty of assessing legal predicates based on facts expressed in the everyday language of discourse.

As reported by \cite{Ashley2009}, the two long-standing goals of AI and law research are the automatic classification of case texts and the forecasting of case outcomes so that law professionals understand them. According to \cite{Sulea2017a}, artificial intelligence ``systems could act as a decision support system or at least a sanity check for law professionals'' \cite[pp. 1]{Sulea2017a}. This finding is of paramount importance to overcome one of the main arguments against the adoption of AI in the field of law: that intelligent systems will eventually replace lawyers. Looking at modern research in this area makes us believe that current researchers seek to help lawyers with the most demanding and repetitive tasks to be better at other tasks where computers cannot help.

However, this task is not trivial. As written in \cite{Surden2014}, many of the tasks undertaken by lawyers appear to require higher-level intellectual skills beyond the capabilities of current AI techniques. However, as highlighted by \cite{Branting2017a}, recent advances in human language technology and large-scale data analysis methods, as well as recent advances in large-scale human language technology, have significantly increased the ability to automate the interpretation of legal text. 

\subsection{Text-based analysis of judicial texts}
In their work, \cite{Aletras2016} have found that formal facts in a case are the most critical predictor factor. This observation is of utmost importance because legal texts also have specific characteristics that differentiate them from other narratives. For example, the use of references in legal texts has specific structures different from the use of references in the public domain, as analyzed by~\cite{Tran2014}. Even before modern times, as \cite{Alarie2017} write, juristic decisions, legislative acts, regulations, and scientific and practice-based commentaries were placed in published volumes to reuse legal texts in various circumstances.

When examining legal precedents via AI, matching a particular case's facts can be difficult. As mentioned by \cite{Zeng2007}, many questions may arise from different perspectives, such as what is relevant law, how to interpret the applicable law in the context, how to apply the law, and the facts of the case. This issue is also stated by \cite{Sannier2017}, who writes that analysts need to consider cross-referencing in legal texts and add information to the cited provisions when identifying and elaborating on legal requirements. This point is also discussed by \cite{Tran2014}, noting that legal texts at the discourse level contain many reference phenomena.

In their work, \cite{Aletras2016} hypothesized that published judgments could be used to test the possibility of text-based analysis for ex-ante outcome forecasting. This idea is corroborated by \cite{Surden2014}, which advocates that entities involved in legal outcomes could probably use past client scenarios and other relevant public or private data to create predictive machine learning. These predictive machines could model future outcomes in specific legal problems that would complement legal counseling.

Moreover, \cite{Liu2017} writes that a valuable way to explain the decisions of past judges and predict future decisions is to study empirical variables that reflect non-legal facts rather than pure legal deductive arguments. In the same thought, \cite{Zeng2007} writes that old legal cases have often been used to support ideas and judicial opinions of law professionals, even without using artificial intelligence or mathematical or statistical models. According to the authors, past cases are called \emph{precedents} in the Common law system and can be followed, analogized, distinguished, or overturned. 

It is important to emphasize that the central core of common law, using precedents, is not a common tenet of civil law, a legal model used in the Brazilian courts. Although precedent usage has only recently become more frequent after the 1988 Brazilian Constitution\footnote{url{http://www.brazil.gov.br/about-brazil/news/2018/11/civil-law-tradition-guides-rights-in-brazil-but-common-law-is-also-present}}, Brazil is still focusing on laws and codes rather than analyzing cases from the past.

\subsection{Recent Works}
Artificial intelligence methods are being used successfully in various areas of law. As an example, \cite{Mcshane2012} has used AI and NLP to build a hierarchical Bayesian model to predict settlements in class action lawsuits of federal securities in the United States. In another work, \cite{Gokhale2017} have developed a co-training algorithm for classifying human rights abuses, using SVM and statistical regression on the domain ontology. 

In the field of topic extraction, \cite{Remmits2017} used Latent Dirichlet Allocation to explore major topic topics of discussion in judicial decisions of the American Supreme Court. In their work, the authors also compare whether legal experts and non-legal people agree with their judgments, finding that domain experts and non-domain experts may assess topics differently. 

Likewise, in legal prediction, our main research objective, recent advances have significantly improved state of the art. In an essential work related to our research, \cite{Aletras2016} used a data set containing cases of the European Court of Human Rights that violated three articles of their Convention. These are:

  \begin{itemize}
      \item Article~3: \emph{Prohibits torture and inhuman and degrading treatment};
      \item Article~6: \emph{Protects the right to a fair trial}; and
      \item Article~8: \emph{Provides a right to respect for one’s private and family life, his home, and his correspondence}.
  \end{itemize}
  
  The authors then captured the same number of cases with violations and non-violations of the three articles mentioned above. After using regular expressions and preprocessing tools to extract text, authors obtained N-gram properties for proceedings, circumstances, facts, relevant law, law, and the entire case itself. After extracting the N-grams, they formed groups using vector space models to find the main topics of each article. They used support vector machines (SVM) to obtain 78\% precision when predicting topics for Article~3, 84\% when predicting topics and events for Article~6, and again 78\% when predicting topics and events for Article~8.

After the prediction step, the authors also examined the weights of their SVMs to find the main words that affected each of the violations. In Article~3, words like ``\emph{injury, protection, ordered, damage, civil, caused, failed, claim, course, connection, region, effective, quashed, claimed, suffered, suspended, carry, compensation, pecuniary, ukraine}'' contributed positively to a violation. On the other hand, ``\emph{sentence, year, life, circumstance, imprisonment, release, set, president, administration, sentenced, term, constitutional, federal, appealed, twenty, convicted, continued, regime, subject, responsible}'' contributed negatively to a violation. The process is repeated for the other two articles.

In another influential study, \cite{Sulea2017a} used lexical characteristics and SVMs to predict the area of law (such as criminal, social or commercial law) and the decisions of the French Supreme Court. The authors used a diachronic suite of French Supreme Court rulings (in French, \emph{Court de Cassation}). The entire collection included 131,830 documents, each with unique decisions and metadata. Standard metadata available in most documents included the law, timestamp, ruling (eg, case management, rejection, non lieu, etc.), case description, and cited laws.

After preprocessing, their data set contained 126,865 different court rulings, each containing a description of the case and four different types of labels. a) The area of jurisdiction; b) The date of the decision; c) Case Decision itself, and; d) List of articles and laws cited within the description. The features are then selected using hierarchical clustering and using SVM to classify the data set. The authors have achieved accuracy as high as 90.2\% to classify the legal field; 96.9\% accuracy using 6-class SVM to classify the court decision, and 74.3\% accuracy using 7-class SVM to determine the date of the case.

These works suggest that applying AI and NLP to law is a growing area that offers new and promising research opportunities. The law is a wide range of applications that generally produce a substantial amount of data, especially text. As a result, researchers can achieve substantial results with relevant applications.

\section{Research Methodology}
\label{meth}
This section describes the characterization of the domain and the necessary steps to complete the research. We also describe evaluation measures that evaluate the effectiveness of the proposed model.

\subsection{Data Collection and Pre-processing}
For our investigation, we collected a corpus of legal decisions from the eSAJ, the judicial system of the TJSP\footnote{\url{http://esaj.tjsp.jus.br/cjpg/}}. We have selected some previously defined legal subjects to limit the captured documents, and we have selected only cases where the judges have very defined results. In our scope, we have second-degree murder cases (\emph{Homicídio simples}), here called homicide, and active corruption cases (\emph{Corrupção Ativa}), here called corruption. We then selected these judicial decisions with the defendant's conviction or absolution. Many different decisions do not have explicit condemnation or absolution terms. Therefore, finding those subjects with clear and established outcomes is paramount.

\begin{figure}
	\centering
 	  \caption{Methodological phases of the research.}
		\includegraphics{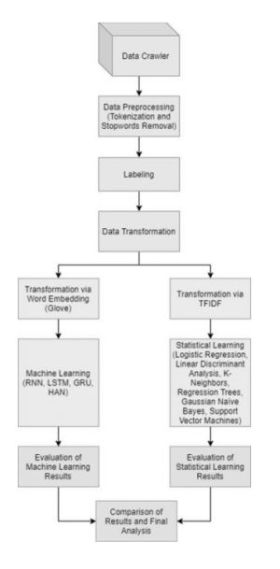}
	\label{fig:preprocessing}
\end{figure}

We implemented a web text crawler to collect data from eSAJ, the São Paulo Court of Justice's electronic system. Since the user can choose from many different areas to display judicial information, such as classes, topics, judges, and process numbers, they can select a specific text and get its full metadata. The crawler stored the documents retrieved from the queries in a file. The Python code for the web crawler is accessible on the author's GitHub page \footnote{\url{https://github.com/vbertalan}}. 

By running the text crawler developed for this study, we collected 2,467 cases, only choosing subjects for homicide and corruption, and 1,681 homicide cases and 786 corruption cases were collected. The crawler was used to collect documents from different periods. The complete distribution from absolutions to condemnations is shown in Table~\ref{totaldocs}. Related crimes did not feature in our scope.

\begin{table}[h!]
\centering
\caption{Distribution of cases for research.}
\label{totaldocs}
\begin{tabular}{|l|r|l|r|l|}
\hline
\textbf{Judicial subject}        & \textbf{Homicide} & \textbf{\%} & \textbf{Corruption} & \textbf{\%} \\ 
\textbf{Number of Absolutions}   & 844               & 50.2        & 197                  & 25.0        \\ 
\textbf{Number of Condemnations} & 837               & 49.7        & 589                 & 75.0        \\ \hline
\textbf{Total Cases}             & 1,681             & 100         & 786                 & 100         \\ \hline
\end{tabular}
\end{table}

The data was preprocessed to remove unnecessary information. The first step was tokenizing words and excluding stopwords, as recommended by \cite{Moens2001}. 
We also labeled the data set to guide the supervised learning algorithms. Each judicial document was classified as condemnation (-1) or absolution (+1). We used professional advice from Brazilian lawyers to better understand the texts. 


As for methods, we start describing two measures used to represent the words in a text, TF-IDF and word embeddings.

\subsection{TF-IDF}
The acronym for `Term Frequency--Inverse Document Frequency' is a numerical statistical measure used to quantify the importance of a word in a document or corpus. Let $N$ be the number of documents in a corpus $D$, $(N=|D|)$. Let also $|\{d \in D: t \in d\}|$ be the number of documents in which the term $t$ appears.
TF-IDF can be calculated according to Equation~\ref{equation_tfidf}, where $\mbox{TF}_{t,d}$ represents the frequency of the term $t$ in $d$, while $\mbox{IDF}_{(t,D)}$ represents the frequency of the inverse document, which is a measure of the amount of information that the word provides.

\begin{align}
\mbox{TF-IDF}_{(t,d, D)} &= \mbox{TF}_{(t,d)} \cdot \mbox{IDF}_{(t,D)} \nonumber \\
                   ~ &= f_{(t,d)}) \cdot \log \frac{N}{n_t} \label{equation_tfidf}
\end{align}

\subsection{Word Embeddings}
We can use various ways to transform data. One of the most popular is word embedding. Word embeddings, as defined by \cite{turian2010}, are vectors composed of real numbers distributed over an interdimensional space induced by semi-supervised learning. Word embeddings have become an effective alternative to convert pure text to mathematical values and reduce data manipulation via machine learning algorithms. The algorithms calculate similarities between two vectors using cosine similarity to calculate similarities between cosine pairs. Each vector dimension represents a characteristic that aims to capture in a distributed manner the semantic, synthetic, or morphological properties of a word.

In this research, we have used GloVe, a method created by \cite{pennington2014glove}. GloVe, a reduction for Global Vectors, is a word embedding method developed to learn word vectors such that their dot product equals the logarithm of the probability of co-occurrence of the words. Rather than establishing a local context through a window, GloVe constructs a clear word context matrix, or word co-occurrence matrix, using a statistical analysis throughout the text corpus to construct the exact word context matrix or word co-occurrence matrix. It can combine a term's local and global representations by mixing features of two family models, namely the global matrix factorization and local context window modeling methods. We used a pre-trained GloVe corpus developed by \cite{ICMCGlove} for Brazilian Portuguese.

\subsection{Machine Learning Algorithms}
We came across different NLP algorithms used in many ways during our bibliography review of the dissertation. We selected algorithms that performed well in similar classification studies so that we could test their precision on our data sets. We show the complete list of algorithms chosen and a few examples of their applications in Table \ref{algorithmchoices}. 

\begin{table}[h!]
\centering
\caption{Chosen algorithms and practical NLP applications.}
\label{algorithmchoices}
\resizebox{\textwidth}{!}
{\begin{tabular}{|c|l|}
\hline
\textbf{Algorithm}                  & \multicolumn{1}{c|}{\textbf{Practical NLP Applications}} \\ \hline
Logistic Regression                 & \vtop{\hbox{\strut \cite{Liu2017}, for prediction of circumstances and topics of law cases}\hbox{\strut \cite{Pelle2018}, for offensive text detection}}           \\ \hline
Linear Discriminant Analysis        &   \vtop{\hbox{\strut \cite{Krestel2009}, for tag recommendation in search websites}\hbox{\strut \cite{Pavlinek2017}, for text classification in newsgroups datasets}}              \\ \hline
K Nearest Neighbors                 & \vtop{\hbox{\strut \cite{Chantar2011}, for document categorization in the Arabic language}\hbox{\strut \cite{Desmet2014}, for automatic recognition of suicidal messages in social media}}                \\ \hline
Classification and Regression Trees &  \vtop{\hbox{\strut \cite{7050801}, for measuring sentiment analysis on Twitter}\hbox{\strut \cite{Rios-Figueroa2011}, for predicting judicial independence in Latin American courts}}               \\ \hline
Naive Bayes                         &  \vtop{\hbox{\strut \cite{Harcourt2015}, for feature selection and vectorization in legal documents}\hbox{\strut \cite{Rios-Figueroa2011}, for measuring sentiment analysis on Facebook statuses}}               \\ \hline
Support Vector Machines             &   \vtop{\hbox{\strut \cite{Do2017}, for legal question answering and ranking}\hbox{\strut \cite{Sulea2017}, for predicting law area and decisions of French Supreme Court cases}}              \\ \hline
Multilayer Perceptron               &   \vtop{\hbox{\strut \cite{Rao2016}, for political text classification}\hbox{\strut \cite{Sa2017}, for defining the author reputation of product comments}}              \\ \hline
Recurrent Neural Networks           & \vtop{\hbox{\strut \cite{Alschner2017}, for automated production of legal texts}\hbox{\strut \cite{Kim2017}, for demographic inference on Twitter}}                 \\ \hline
Long Short Term Memory      &  \vtop{\hbox{\strut \cite{Li2017}, for political ideology analysis}\hbox{\strut \cite{Xie2017}, for mining product adverse events in social media}}               \\ \hline
Gated Recurring Unit        &  \vtop{\hbox{\strut \cite{Luo2017}, for predicting charges for criminal cases}\hbox{\strut \cite{zhang2018detecting}, for detecting hate speech on Twitter}}               \\ \hline
Hierarchical Attention Networks     & \vtop{\hbox{\strut \cite{Branting2017a}, for predicting models for decision support in administrative adjudication
}\hbox{\strut \cite{gao2018hierarchical}, for information extraction from cancer pathology reports}}                \\ \hline
\end{tabular}}
\end{table}

\subsubsection{Logistic Regression}

Logistic regression is a model used to predict categorical variables from a series of explanatory continuous or binary variables. Like all regression analyzes, logistic regression is a predictive analysis, and it is used to predict the occurrence of an event directly. The algorithm takes a weighted combination of input features (in our study, the result of the TFIDF transformation) and weights it based on the combination of input features. The result passed through a sigmoid function which converts an actual float number to a numeric value between 0 and 1.

As cited by \cite{10.5555/1162264}, it operates as a statistical method to find equations that predict the outcome of a binary variable from one or more response variables. Since the model does not strictly require continuous data, the response variables can be categorical or continuous. Logistic regression uses logarithmic odds ratios rather than probabilities to predict group membership and iterative maximum likelihood methods instead of least squares to fit a final model. 

\subsubsection{Linear Discriminant Analysis}

Linear discriminant analysis is a multivariate statistical method for discriminating and classifying objects. Each sample is classified into one of many populations with a \emph{p} number of features, thus minimizing the likelihood of incorrect classification. To this end, the algorithm uses a combination of linear features that exhibit a higher classification capability between populations to be able to do so. This combination is called a discriminant function.

In their work, \cite{balakrishnama1998linear} argue that linear discriminant analysis considers situations where within-class frequencies are not equally distributed. Their performance was tested against randomly generated test data. A discriminant function is designed to confirm which variable of the subset is essential to classify this subset among populations. Since our study aims to rank the subset between two groups of condemnation and absolution, we used a linear discriminant analysis method. This approach uses a discriminant function to classify each value among two populations by selecting the minimum ratio of the difference between the pairs of means of the multivariate group and the variance of the multivariate within two groups.

\subsubsection{K-Nearest Neighbors - KNN}
The K-Nearest Neighbors (KNN) algorithm is a supervised learning algorithm that aims to find \emph{k}-labeled examples closest to non-classified examples by labeling the most comparable examples. Algorithms in the KNN family do not require significant computational effort during training. However, the computational cost for labeling a new instance is significantly high because, in the worst-case scenario, all other instances present in the training data set must be compared with this example.

As stated by \cite{10.5555/1162264}, in high data density regions, models can lead to over-smoothing and cleaning a structure that could otherwise be extracted from a data structure. However, simplifying the model can lead to noisy estimates in other parts of the data space where the density is more diminutive. Therefore, optimal model selection can depend on its location in the data space. The approach of nearest neighbor density estimates addresses this problem.

\subsubsection{Regression Trees}
Classification and Regression Trees (CARTs) are machine learning methods for building data-based prediction models. As stated by \cite{lohtrees}, models are obtained by recursively splitting the data space and fitting a simple prediction model in each partition. Classification trees employ dependent variables that take a finite number of unordered values, with prediction errors measured in misclassification costs as the prediction error. Regression trees are for dependent variables taking continuous or ordered discrete values, where the square difference between the observed values and predicted values usually measures prediction error.

In this investigation, we use the CART version available in the NLTK framework. The analysis method uses classification rules made by decision trees and begins with the root node with all text properties. The next node contains subsets and subsets of the data, and each division results in precisely two nodes. This method allows for the identification of uniform data sets by systematically comparing their characteristics to establish a relationship between the explanatory variables and one answer variable: in our case, the label of condemnation or absolution. Successive divisions adjust the model in the data set, and the subsets are made to be more homogeneous concerning the answer variable with each successive division. The division process is repeated until either none of the selected variables shows a significant effect in the division, or the subset is too small to be split again.

\subsubsection{Naïve Bayes}

The Naïve Bayes classifier is a family of probabilistic classifiers that use Bayes's theorem to generate models with high assumptions of independence between features. As written in \cite{Manning:2008:IIR:1394399}, Bayesian classifiers are those in which an object $x$ is assigned to the class $C_{k}$ based on the probability of $x$ belonging to $C_{k}$. We present an example of the formula in Equation \ref{equationbayes}.

\begin{equation}
\label{equationbayes}
\mbox{P}(C_{k} | x) = \frac{\mbox{P}(C_{k})\mbox{P}(x | C_{k})}{\mbox{P}(x)}
\end{equation}
where:
\begin{itemize}
	\item ${P}(C_{k} | x)$ is the probability of hypothesis $C_{k}$ given the data $x$. This value is called posterior probability.
	\item ${P}(C_{k})$ is the probability that hypothesis $C_{k}$ is true (irrespective of the data). This value is called the prior probability of $C_{k}$.
	\item ${P}(x | C_{k})$ is the probability of data $x$ given that the hypothesis $C_{k}$ was true. 
	\item ${P}(x)$ is the probability of the data $x$ (irrespective of hypothesis).
\end{itemize}

\subsubsection{Support Vector Machines}
Support vector machines (SVM) are an algorithm for binary classification that plots elements of a dataset and attempts to divide them by defining a separation function. The most effective separation function is that that shows the best classification by providing the most margin between the two given classes. In this model, the support vectors are dots of both classes closest to the separation function in the center of the support vectors. This separation function is also known as a hyperplane. The algorithm plots the new element in the same space to predict new features and ensures the appropriate grouping of new elements.

As stated in \cite{10.5555/1162264}, if multiple solutions precisely classify training data sets, we should try to find one that gives the slightest generalization error. According to the author, support vector machines approach this problem through the concept of margin. This concept is defined as the smallest distance between the boundary of the decision and one of the samples. In support vector machines, the decision limit is selected to maximize the margin.

\subsubsection{Multilayer Perceptrons - MLP}

Multilayer perceptrons are algorithms that extract features from a data set consisting of interconnected units called \emph{perceptron neurons}. These neurons are units responsible for controlling the errors that result from each algorithm-predictive projection.

The MLP architecture operates in layers, and the first layer functions as a sensory receptor that receives an input data signal. The last layer is called the exit layer, and in this layer we can see the response of MLP to the input signal. In between the first layer and the exit layer, we can have some hidden layers. The hidden and exit layers are perceptron neurons that receive the input signal, process the signal by activation functions, and pass the signal to the next layer. 

The weights (synapses) must be calibrated after each input to allow the network to learn the essential characteristics of the data set. These synapses are calibrated using an algorithm intended to minimize MLP errors, known as \emph{backpropagation}. 

\subsubsection{Recurrent Neural Networks - RNN}
MLPs work for many applications. However, if we have inputs that behave like time series when the value is intrinsically dependent on previous outputs, this signal will directly affect the following input. Values tend to fall out inside the MLP architecture since all inputs receive singular outputs without dependency between them.

Recurrent neural networks are an architecture that handles sequential inputs of variable length using sequential shared hidden states. An RNN is a machine learning model with $M$ inputs fully connected to $N$ units. Since the meaning of words in a text depends entirely on the previous and post-progressive terms, we can also consider the text as time series and powerful candidate for RNNs.

In RNN the output of a unit in the $n+1$ step depends not only on the network exits of the previous step $(u(n-k),k=0,...,M-1)$, but also on the previous outputs of the units $y_k(n),k=1,...,N$. Feedback inputs are provided in the recurring layer, allowing the network to store information over epochs in memory.

\subsubsection{Long Short Term Memory Networks -- LSTM}

After running an RNN for several epochs, a known problem that can occur after running the RNN is called the \emph{vanishing gradient} problem, in which the gradient of the loss function changes exponentially over time, effectively preventing the weight from changing its value. In this scenario, RNN fails and does not offer valuable prediction capability.

Long-term short-term memory networks (LSTM) are different types of RNN architectures aimed at avoiding the vanishing gradient problem. In addition to standard units, LSTM networks use select units, and these units add memory cells that can store data in memory longer than ordinary RNNs. 

A set of gates is used to control whether and when data enter the memory (\emph{input gate}), whether an output is emitted (\emph{output gate}), and when the previous data are forgotten during the next epoch ( \emph{forget gate}). This RNN architecture allows the network to store only valuable information in subsequent epochs and effectively learn dependencies over extended periods, preserving valuable information for subsequent epochs (\emph{Long Term}). Information that does not add to the model is discarded (\emph{Short Term}). 

\subsubsection{Gated Recurring Unit Networks - GRU}
Gated recurring network networks are based on the LSTM architecture with notable differences, such as the absence of memory cells and a gate output. As an alternative to these exclusions, the GRU architecture operates a \emph{reset gate} and a \emph{update gate}. The reset gate works by comparing the previous activation with the subsequent candidate activation to discard/forget previous states and decide whether the candidate activation will be used to update the cell state. 

While LSTMs limit the state of cells by controlling their gates, GRUs expose memory content to other architecture units. Without restrictions and with a simpler model, GRUs generally train faster than LSTMs without limitations.

\subsubsection{Hierarchical Attention Networks -- HAN}
Hierarchical Attention Networks, described by \cite{yang-etal-2016-hierarchical}, is a neural network architecture highlighting the importance of individual words or sentences in document representations. Since not all words are equally crucial in text classification and sentences do not all have the same meaning, this model emphasizes the most important sequences that affect the class/label of the document.

HANs are usually made up of 6 layers:

\begin{itemize}
	\item An \emph{embedding layer}, the layer responsible for creating a matrix with the characteristics (size of the vocabulary, maximum length of sentences) of the documents being processed 
	\item A \emph{word sequence encoder}, a bidirectional word level GRU to obtain a rich representation of words
	\item A \emph{word attention layer}, a layer to obtain important information in a sentence
	\item A \emph{sentence encoder}, a bidirectional sentence level GRU to get a rich representation of words
	\item A \emph{sentence attention layer}, a layer to obtain important information in a sentence
	\item A \emph{final layer}, destined to fully connect all the previous output and apply a softmax activation function.
\end{itemize}

\begin{figure}
	\centering
 	  \caption[Diagrammatic example of a HAN.]{Diagrammatic example of a HAN. Source: \cite{yang-etal-2016-hierarchical}.}
		\includegraphics{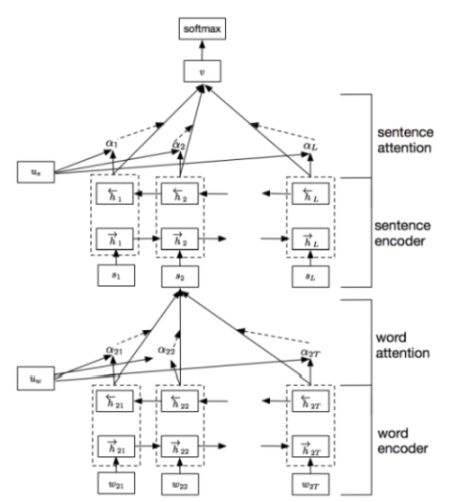}
	\label{fig:han}
\end{figure}

HAN works by queuing a substructure called the word encoder and another substructure called the sentence encoder. The first applies attention to each word inputted to create sentence representations, and the second adds attention to each sentence received before creating document representations. Figure \ref{fig:han} shows an example of a HAN.  

For \textbf{word encoder}, the following structure was built:

\begin{itemize}
	\item An \textbf{input layer}, to receive GloVe's output, the tokens $w_{i_t}$, representing the word $i$ per sentence $t$, in a matrix of None per N, N representing the maximum words in a sentence in the data set. It is essential to mention that the term \emph{None} is used by Keras, the Python framework that we have used, to represent any scalar number so that we can use this model to infer on an arbitrarily long input. 
	\item To make the model understand sequences of characters, we have an \textbf{embedding layer}, destined to process strings. This layer assigns multidimensional vectors $W_{ew_{i_j}}$ to each token. Therefore, words are represented numerically as $x_{i_t}$, as a projection of the term in a continuous vector space. There are many embedding methods available. For this research, we used the GloVe framework. This layer will output a matrix with None per N per the number of dimensions of the word embedding training file, in our case, 600. 
	
 	\item The third layer contains an \textbf{encoding layer}, in our case a bidirectional GRU, to encode the data. Bidirectionality works by reading the sentence from the first to the last word and reversing the order afterward to understand the connections between words on the left and the right. As an example, in the sentence \emph{The black car is beautiful}, the term $black$ is directly related to the word $car$, as it gives a character to the word, and the word $is$ also represents a strong correlation with $car$, indicating that the following word will describe its nature. The context annotations outputted are represented by $h_{i_t}$.
 	
	\item The following \textbf{dense layer} works by applying the activation function (in our case, ReLU, to counteract the problem of the vanishing gradient) to return the neural network's output. 
	
	\item Subsequently, the result is processed in \textbf{word attention layer}, which is an MLP destined to learn the importance of words through training with randomly initialized weights ($W$), biases ($b$), and the output of the encoding layer, as in Equation \ref{eqhan1}:
	\begin{equation}
	\label{eqhan1}
    u_{i_t} = tanh(W_w h_{i_t} + b_w)
    \end{equation}

    After that step, the result $u_{i_t}$ is then multiplied by a trainable context vector $u_w$ and normalized to an importance weight per word $\alpha_{i_t}$ by a softmax function, described in Equation \ref{eqhan2}. The word context vector $u_w$ is randomly initialized and jointly learned during the training process.
    
    \begin{equation}
    \label{eqhan2}
    \alpha_{i_t} = \frac{\exp(u_{i_t}^T u_w)}{\sum_{t}^{}\exp(u_{i_t}^T u_w)}
    \end{equation}

Finally, those importance weights $\alpha_{i_t}$ are multiplied by context annotations $h_{i_t}$, called sentence vectors, and entered into the sentence encoder. This operation is described in Equation \ref{eqhan3}. 
    
    \begin{equation}
    \label{eqhan3}
    s_i = \sum_{t}^{}\alpha_{i_t} h_{i_t}
    \end{equation}
\end{itemize}

For \textbf{sentence encoder}, the following structure was built: 

\begin{itemize}
	\item The \textbf{input layer} receives the result from the last layer of word attention, with a matrix of None per M per N, where M is the maximum number of sentences in one document and N is the maximum number of words in a sentence in the data set.
	\item The second layer represents \textbf{time distributed model}, which is responsible for wrapping every input it receives as a dense layer, applied to all layer layers at the word level in each sentence. In contrast, a regular dense layer would compute all the inputs as single N units.
	\item As in the word encoder, an \textbf{encoding layer} is used, in our case, a Bidirectional GRU. As mentioned above, the GRU is used to understand the semantic relations between the sentences. 
	\item Then, a \textbf{dense layer} is stacked, with ReLu activation, to retrieve the outputs $h_i$. 
	\item Finally, the result is entered into the \textbf{sentence attention layer}. It works similarly to the word attention layer, but the final output is a document vector $v$, which can be used as a feature for document classification. Trainable weights and biases are again randomly initialized and jointly learned during the training process. The operation is described in Equations \ref{eqhan4}, \ref{eqhan5}, and \ref{eqhan6}.  
	
		\begin{equation}
		\label{eqhan4}
    u_{i_t} = tanh(W_s h_i + b_s)
    \end{equation}
    
        \begin{equation}
        \label{eqhan5}
    \alpha_{i_t} = \frac{\exp(u_i^T u_s)}{\sum_{t}^{}\exp(u_i^T u_s)}
    \end{equation}
    
        \begin{equation}
        \label{eqhan6}
    v = \sum_{i}^{}\alpha_i h_i
    \end{equation}
\end{itemize}

\begin{figure}[h!]
	\centering
 	  \caption[Example of attention generated by the HAN.]{Example of attention generated by the HAN. Source: \cite{yang-etal-2016-hierarchical}.}
		\includegraphics{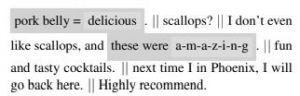}
	\label{fig:exampleattentionhan}
\end{figure}

After processing the HAN networks, every word gets an attention coefficient, indicating the importance of that word in its sentence. An example of \cite{yang-etal-2016-hierarchical} is shown in Figure \ref{fig:exampleattentionhan}. It can see that sentence 1 (\emph{"pork belly = delicious"}) and the final words of sentence 3 (\emph{"these were a-m-a-z-i-n-g"}) are marked in pink. This highlight happens because the HAN has implied that those two sentences are among the most important of that text; that is, they are among the highest sentence attention weights. Inside those sentences, two words are marked in blue. This means those two words carry essential terms in those sentences and have the highest word attention weights.

HANs are currently one of the most popular neural network algorithms being adopted in Computer Science. In 2022, the year of publication of this research, we have seen their applications in many different fields, such as biomedicine \citep{dar2022spectral}, psychology \citep{tamilarasan2022early}, biology \citep{he2022hierarchical}, and recommendation systems \citep{roy2022optimal}.

The adoption of HANs in the legal field is a novel approach that has been positively adopted worldwide. In the paper by Chalkidis~\cite{chalkidis2019extreme} HANs are also used to classify the field of legal texts, and, as previously mentioned, \cite{ma-etal-2019-sentence}, used the method to predict the result of claim verifications. Variations of the HAN model can be seen in \cite{liu2019}. In their paper, they created a variation of the HAN model to determine charges in criminal cases or types of dispute in civil cases according to the fact descriptions. In \cite{WANGWenguan505}, they adopted an improved version of the algorithm for predicting crime, recommending legal articles, and predicting sentences from judicial documents. In \cite{chi2022legal}, they used HANs to predict legal judgments in commutation cases. 

Our research is the first, to the best of our knowledge, to predict judicial outcomes in Brazilian Portuguese, with results that emulate those found in different languages in the field of law. 

\section{Experiments}
\label{exp}
In this chapter, we present the experiments and results after processing the data sets for each of the algorithms selected, both neural and non-neural. We also show the word attention weights for each data set and for each word. Finally, we show the top-ranked words in all scenarios.

We emphasize that several different hyperparameter combinations were tried before the optimal match was found. The framework-provided standard set was the best option in all cases where we did not mention the hyperparameters used.

\subsection{Non-Neural Networks}
We used K-fold cross-validation with 10 folds for all methods in this section. After conducting our tests, we concluded that regression trees were the method that had the highest accuracy in the chosen data sets, even though the other algorithms showed good performance. For example, SVM showed good performance in the homicides dataset but did not correspond to the performance regression trees shown in the corruption dataset. Regression trees have always managed to predict good results. 

These results echo other results found in other studies in the legal field in many different countries, such as the work of \cite{Kastellec2010}, who obtained good results using regression trees in the US legal system. The author mentions that regression trees can examine the inherent conceptions of law to reveal patterns that other methods cannot replicate as effectively.

Other researchers have also used the same method, such as \cite{Rios-Figueroa2011}, who used regression trees to analyze the concepts of justice independence and corruption among supreme courts in Latin America. \cite{Antonucci2014} adopted regression trees to measure efficiency in Italian courts; and \cite{Kufandirimbwa2012}, used the same algorithm to predict results in Zimbabwe.

These studies show that although the world's legal system and other languages and countries such as Brazil, the United States, Italy, and Zimbabwe differ entirely, they have similarities that can be measured effectively with appropriate algorithms. In this way, legal texts can have intrinsic factors that may remain even if languages change.

\subsection{Homicides Dataset}
The metrics for the homicide data set after the experiments are shown in Table \ref{metrics_algorithms_homicides}. Logistic regression, linear discriminant analysis, and support vector machine showed the highest performance with high accuracy, recall, f-score, and precision values. Support vector machines showed the best performance in 3 out of the 4 metrics, and the regression tree showed the highest recall value.

\begin{table}[h!]
\centering
\caption{Metrics of the algorithms, in the homicides dataset, with mean values after 10-fold cross-validation.}
\label{metrics_algorithms_homicides}
\begin{tabular}{|l|l|l|l|l|}
\hline
\textbf{Algorithm}           & \textbf{Precision} & \textbf{Recall} & \textbf{F-Score} & \textbf{Accuracy} \\ \hline
Logistic Regression          & 0.948895        & 0.934847     & 0.939733      & 0.941117       \\ \hline
Linear Discriminant Analysis & 0.921595        & 0.928063      & 0.922120      & 0.923271       \\ \hline
K-Neighbors                  & 0.779389        & 0.820864     & 0.795847      & 0.795330       \\ \hline
Regression Trees             & 0.888953        & \textbf{0.954632}      & 0.888741      & 0.892924       \\ \hline
Naive Bayes                  & 0.651370        & 0.894028     & 0.769831      & 0.723989       \\ \hline
Support Vector Machines      & \textbf{0.951587}        & 0.933694     & \textbf{0.940827}      & 
\textbf{0.952380}       \\ \hline
\end{tabular}
\end{table}

\subsection{Corruption dataset}
\label{corruptiondataset}
The metrics of the corruption data set after the first experiment are shown in table \ref{metrics_algorithms_corruption} after the conclusions of our experiments. We can see that the regression trees were the best algorithm and scored the highest on all four metrics.

\begin{table}[h!]
\centering
\caption{Metrics of the algorithms, in the corruption dataset, with mean values after a 10-fold cross-validation.}
\label{metrics_algorithms_corruption}
\begin{tabular}{|l|l|l|l|l|}
\hline
\textbf{Algorithm}           & \textbf{Precision}   & \textbf{Recall}      & \textbf{F-Score}     & \textbf{Accuracy}    \\ \hline
Logistic Regression          & 0.814269                  & 0.827583                  & 0.870421                  & 0.824269          \\ \hline
Linear Discriminant Analysis & 0.859783                 & 0.828997          & 0.807658          & 0.839766          \\ \hline
K-Neighbors                  & 0.851981                  & 0.828655          & 0.897243          & 0.824854          \\ \hline
Regression Trees             & \textbf{0.967917} & \textbf{0.968705} & \textbf{0.973648} & \textbf{0.968421} \\ \hline
Naive Bayes                  & 0.876901                  & 0.899474          & 0.925169          & 0.866374          \\ \hline
Support Vector Machines      & 0.876901                  & 0.872200          & 0.930724         & 0.876901          \\ \hline
\end{tabular}
\end{table}

\subsection{Neural Networks}
\label{neural}
We used a K-fold cross-validation for experiments in neural networks with ten folds. The same methodology applies to the homicide and corruption data sets. All tests were processed with the GloVe file destined for Brazilian Portuguese with 600 embedding dimensions.

Convergence criteria were defined by modeling learning rate and loss function models. The learning rate is lowered by 0.2 every three epochs, so the loss function remains stable. In the loss function, the algorithm stops after five epochs without a reduction of at least 0.001. 

The results of the homicide data set can be seen in Table \ref{metrics_homicides}. The results of the corruption data set are shown in Table \ref{metrics_corruption}. 

\begin{table}[h!]
\centering
\caption{Metrics of neural networks on the homicide dataset, with mean values after a 10-fold validation}
\label{metrics_homicides}
\begin{tabular}{|l|l|l|l|l|}
\hline
\textbf{Dataset} & \textbf{Precision} & \textbf{Recall}   & \textbf{F-Score}  & \textbf{Accuracy} \\ \hline
MLP              & 0.981267           & 0.986228          & 0.983620          & 0.984562          \\ \hline
RNN              & 0.864426           & 0.882808          & 0.866621          & 0.853257          \\ \hline
LSTM             & 0.986512           & 0.985387          & 0.985890          & 0.986355          \\ \hline
GRU              & \textbf{0.992026}  & \textbf{0.993282} & \textbf{0.992625} & \textbf{0.992275} \\ \hline
HAN              & 0.966543           & 0.986666          & 0.976097          & 0.986666          \\ \hline
\end{tabular}
\end{table}

\begin{table}[H]
\centering
\caption{Metrics of neural networks on the corruption dataset, with mean values after a 10-Fold validation.}
\label{metrics_corruption}
\begin{tabular}{|l|l|l|l|l|}
\hline
\textbf{Dataset} & \textbf{Precision} & \textbf{Recall}   & \textbf{F-Score}  & \textbf{Accuracy} \\ \hline
MLP              & 0.749448           & 0.987332          & 0.855894          & 0.749448          \\ \hline
RNN              & 0.797958           & 0.989252          & 0.882732          & 0.804203          \\ \hline
LSTM             & 0.940756           & 0.986466          & 0.962574          & 0.942762          \\ \hline
GRU              & \textbf{0.996551}  & \textbf{0.998245} & \textbf{0.997391} & 0.996551          \\ \hline
HAN              & 0.985882           & 0.993251          & 0.985540          & \textbf{0.997853} \\ \hline
\end{tabular}
\end{table}

We can see that we got the highest metrics using GRUs and HANs. For the former, the accuracy gains shown for previous models for the GRUs are maintained, and errors are reduced, even for simplified structures with two gates (reset and update gates) on the GRUs. An important point to note is that, in all of our experiments, the GRU training was faster than the LSTM in terms of efficiency. Therefore, we have higher accuracy values and faster training times, making GRUs a good choice for handling judicial texts.

Our results match those found by \cite{chung2014empirical}, which showed that GRU is faster than LSTM, but has a comparable accuracy. The authors also mentioned that choosing a network type between LSTMs and GRUs can depend heavily on the database and the corresponding task. Our studies show that for our datasets, GRU offers higher metrics. Regarding HAN, our results are similar to other work comparing these network structures with other neural network models found in academia, such as HAN from \cite{gao2018hierarchical}, who used HAN to extract information from pathological cancer reports. 

The HAN architecture showed promising results comparable to those found with the GRU, but it has a longer training time. However, both algorithms proved to be high-accuracy choices for our datasets.GRUs have high accuracy but do not implement the Attention model and do not give Attention Weights to each word token. If outcome prediction is the only interest of a study, GRU may be a choice with high accuracy. However, HANs can be used without a significant loss of overall accuracy if an analysis of word tokens is required.

\section{Analysis of Attention Weights}
\label{attention}
We calculated all attention weights for each word in the data sets. Because words have different semantic meanings in different sentences, the occurrence of these words in our attention weight dataset has been repeated multiple times.

Since absolution and condemnation are different document classifications, words can also have different attention weights in both cases. This difference is why we mapped each of the datasets twice for each result to map all possible words that influence the meaning of each sentence. Table \ref{number_tokens} shows the total number of word tokens. 

\begin{table}[H]
\centering
\caption{Number of word tokens for each data set}
\label{number_tokens}
\begin{tabular}{|l|l|}
\hline
\textbf{Dataset}         & \textbf{Number of Unique Tokens} \\ \hline
Homicide Absolutions     & 248460                           \\ \hline
Homicide Condemnations   & 466461                           \\ \hline
Corruption Absolutions   & 66929                            \\ \hline
Corruption Condemnations & 252620                           \\ \hline
\end{tabular}
\end{table}

After the classification was performed, we tried to organize all the words in each data set according to their attention weights. Therefore, each word has a unique value from 0 (where the word would not be important in document classification) to 1 (where the word would be of the highest importance in document classification).

Note that a word can be assigned different attention weights in different sentences. As a short example, the sentence \emph{The defendant robbed a bank} and the sentence \emph{The defendant did not participate in the robbery, because it was going to a blood bank} both have the word \emph{bank}, but in very different contexts. The word in the first sentence contributes significantly to condemnation, while the word in the second contributes greatly to absolution.

Therefore, in our final calculations, words appeared more than once with different attention weights. The list with the top 20 words for the corruption data set and the top 50 words for the homicide data set is listed below in Tables \ref{waw_corruption} and \ref{waw_homicide}, respectively

\begin{table}[]
\centering
\caption{Top 20 word attention weights for the corruption dataset and their English translations}
\label{waw_corruption}
\begin{tabular}{|l|l|l|l|l|l|}
\hline
\multicolumn{3}{|c|}{\textbf{Corruption Absolutions}}         & \multicolumn{3}{c|}{\textbf{Corruption Condemnations}}        \\ \hline
\textbf{Rank} & \textbf{Word} & \textbf{Weight} & \textbf{Rank} & \textbf{Word} & \textbf{Weight} \\ \hline
1                 & real    \emph{(real)}        & 0.98                      & 1                 & assunto    \emph{(subject)}     & 0.99                      \\ \hline
2                 & irregular    \emph{(irregular)}   & 0.97                      & 2                 & supra    \emph{(above)}       & 0.99                      \\ \hline
3                 & nenhuma     \emph{(none)}     & 0.96                      & 3                 & autos     \emph{(case files)}       & 0.98                      \\ \hline
4                 & àquela   \emph{(that one)}        & 0.95                      & 4                 & apresentou   \emph{(presented)}    & 0.98                      \\ \hline
5                 & ofereceu    \emph{(offered)}     & 0.94                      & 5                 & decisão   \emph{(decision)}       & 0.98                      \\ \hline
6                 & reconheceu   \emph{(recognized)}    & 0.94                      & 6                 & cpp    \emph{(criminal code)}          & 0.98                      \\ \hline
7                 & apenas  \emph{(just)}       & 0.94                      & 7                 & decisão   \emph{(decision)}     & 0.97                      \\ \hline
8                 & pública   \emph{(public)}     & 0.93                      & 8                 & público   \emph{(public)}     & 0.97                      \\ \hline
9                 & levado  \emph{(taken)}       & 0.93                      & 9                 & regime   \emph{(regime)}      & 0.97                      \\ \hline
10                & parcialmente \emph{(parcially)}  & 0.93                      & 10                & valdefran  \emph{(first name)}    & 0.97                      \\ \hline
11                & memoriais  \emph{(memorials)}    & 0.92                      & 11                & contou   \emph{(told)}      & 0.96                      \\ \hline
12                & resta  \emph{(remains)}        & 0.92                      & 12                & demonstrada \emph{(demonstrated)}   & 0.96                      \\ \hline
13                & multa    \emph{(fee)}      & 0.92                      & 13                & ativa  \emph{(active)}        & 0.96                      \\ \hline
14                & localidade  \emph{(location)}   & 0.91                      & 14                & exame \emph{(exam)} & 0.96                      \\ \hline
15                & originário   \emph{(originary)}  & 0.91                      & 15                & ministério  \emph{(ministry)}  & 0.96                      \\ \hline
16                & inicial   \emph{(initial)}    & 0.90                      & 16                & começou \emph{(started)}      & 0.96                      \\ \hline
17                & segue    \emph{(follows)}     & 0.89                      & 17                & quantia   \emph{(amount)}    & 0.96                      \\ \hline
18                & oferecendo \emph{(offering)}   & 0.89                      & 18                & propina   \emph{(bribe)}    & 0.96                      \\ \hline
19                & polícia    \emph{(police)}   & 0.88                      & 19                & polícia   \emph{(police)}    & 0.96                      \\ \hline
20                & pretensão   \emph{(pretense)}    & 0.88                      & 20                & peculato  \emph{(embezzlement)}   & 0.96                      \\ \hline
\end{tabular}
\end{table}

\begin{table}[]
\centering
\caption{Top 50 word attention weights for the homicide dataset and their English translations}
\label{waw_homicide}
\begin{tabular}{|l|l|l|l|l|l|}
\hline
\multicolumn{3}{|c|}{\textbf{Homicide Absolutions}}           & \multicolumn{3}{c|}{\textbf{Homicide Condemnations}}          \\ \hline
\textbf{Rank} & \textbf{Word} & \textbf{Weight} & \textbf{Rank} & \textbf{Word} & \textbf{Weight} \\ \hline
1                                  & bo \emph{(incident report)}            & 0.521                            & 1                                  & bo \emph{(incident report)}            & 0.521                            \\
2                                  & mogi \emph{(Brazilian city)}           & 0.428                            & 2                                  & cristina \emph{(Brazlian name)}        & 0.492                            \\
3                                  & comarca   \emph{(county)}              & 0.417                            & 3                                  & horário   \emph{(time)}                & 0.479                            \\
4                                  & estado \emph{(state)}                  & 0.416                            & 4                                  & infração \emph{(infraction)}           & 0.464                            \\
5                                  & santos   \emph{(Brazilian city)}       & 0.414                            & 5                                  & penal   \emph{(criminal)}              & 0.456                            \\
6                                  & justiça \emph{(justice)}               & 0.413                            & 6                                  & cf \emph{(Federal Constitution)}       & 0.452                            \\
7                                  & bem   \emph{(good)}                    & 0.411                            & 7                                  & regime   \emph{(regime)}               & 0.445                            \\
8                                  & sala \emph{(room)}                     & 0.407                            & 8                                  & xavier \emph{(Brazilian name)}         & 0.444                            \\
9                                  & cep   \emph{(postal code)}             & 0.407                            & 9                                  & homicídio   \emph{(homicide)}          & 0.442                            \\
10                                 & competência \emph{(competence)}        & 0.403                            & 10                                 & qualificado \emph{(aggravated)}        & 0.442                            \\
11                                 & antes \emph{(before)}                  & 0.389                            & 11                                 & disparos \emph{(gun shots)}            & 0.440                            \\
12                                 & volta \emph{(return)}                  & 0.387                            & 12                                 & sant \emph{(unknown token)}            & 0.438                            \\
13                                 & origem   \emph{(origin)}               & 0.387                            & 13                                 & exposto   \emph{(exposed)}             & 0.430                            \\
14                                 & infância \emph{(childhood)}            & 0.379                            & 14                                 & provisório \emph{(provisory)}          & 0.424                            \\
15                                 & social   \emph{(social)}               & 0.373                            & 15                                 & mediante   \emph{(through)}            & 0.422                            \\
16                                 & porque \emph{(why)}                    & 0.369                            & 16                                 & philipe \emph{(Brazilian name)}        & 0.421                            \\
17                                 & júri   \emph{(jury)}                   & 0.363                            & 17                                 & sentença   \emph{(sentence)}           & 0.419                            \\
18                                 & principal \emph{(main)}                & 0.352                            & 18                                 & toledo \emph{(Brazilian name)}         & 0.417                            \\
19                                 & júri   \emph{(jury)}                   & 0.352                            & 19                                 & osmarina   \emph{(Brazilian name)}     & 0.416                            \\
20                                 & altura \emph{(height)}                 & 0.349                            & 20                                 & juízo \emph{(in court)}                & 0.415                            \\ 
21                                 & placas \emph{(signs)}                  & 0.349                            & 21                                 & ip \emph{(police investigation)}       & 0.415                            \\
22                                 & nunes \emph{(Braziilan name)}          & 0.348                            & 22                                 & narra \emph{(tells)}                   & 0.413                            \\
23                                 & p   \emph{(page)}                      & 0.348                            & 23                                 & golpes   \emph{(blows)}                & 0.409                            \\
24                                 & machado \emph{(Brazilian name)}                   & 0.346                            & 24                                 & justiça \emph{(justice)}               & 0.409                            \\
25                                 & porte   \emph{(weapon carry)}          & 0.344                            & 25                                 & sala   \emph{(room)}                   & 0.407                            \\
26                                 & agnaldo \emph{(Brazilian name)}        & 0.343                            & 26                                 & acusação \emph{(accusation)}           & 0.403                            \\
27                                 & sp   \emph{(Brazilian state)}          & 0.338                            & 27                                 & sentença   \emph{(sentence)}           & 0.401                            \\
28                                 & anos \emph{(years)}                    & 0.335                            & 28                                 & marta \emph{(Brazilian name)}          & 0.400                            \\
29                                 & regina   \emph{(Brazilian name)}       & 0.332                            & 29                                 & estado   \emph{(state)}                & 0.398                            \\
30                                 & tititi \emph{(gossip)}                 & 0.328                            & 30                                 & silva \emph{(Brazilian name)}          & 0.394                            \\
31                                 & permitido \emph{(allowed)}             & 0.327                            & 31                                 & estado \emph{(state)}                  & 0.393                            \\
32                                 & cor \emph{(color)}                     & 0.326                            & 32                                 & sassolli \emph{(Brazilian name)}       & 0.391                            \\
33                                 & mãe   \emph{(mother)}                  & 0.325                            & 33                                 & prisão   \emph{(prison)}               & 0.389                            \\
34                                 & josé \emph{(Brazilian name)}           & 0.322                            & 34                                 & rua \emph{(street)}                    & 0.387                            \\
35                                 & instrução   \emph{(instruction)}       & 0.322                            & 35                                 & sentença   \emph{(sentence)}           & 0.386                            \\
36                                 & cento \emph{(cent)}                    & 0.322                            & 36                                 & justiça \emph{(justice)}               & 0.385                            \\
37                                 & comum   \emph{(common)}                & 0.321                            & 37                                 & socos   \emph{(punches)}               & 0.385                            \\
38                                 & réu \emph{(defendant)}                 & 0.320                            & 38                                 & análise \emph{(analysis)}              & 0.384                            \\
39                                 & cosmópolis   \emph{(Brazlian city)}    & 0.319                            & 39                                 & flores   \emph{(flowers)}              & 0.382                            \\
40                                 & estado \emph{(state)}                  & 0.319                            & 40                                 & estrita \emph{(strict)}                & 0.382                            \\
41                                 & saído \emph{(gone)}                    & 0.318                            & 41                                 & mínimo \emph{(minimum)}                & 0.382                            \\
42                                 & soubessem \emph{(if they knew)}        & 0.317                            & 42                                 & competência \emph{(competence)}        & 0.381                            \\
43                                 & ação   \emph{(action)}                 & 0.317                            & 43                                 & lesões   \emph{(lesions)}              & 0.379                            \\
44                                 & tribunal \emph{(court)}                & 0.314                            & 44                                 & infância \emph{(childhood)}            & 0.379                            \\
45                                 & pública   \emph{(public)}              & 0.313                            & 45                                 & artigos   \emph{(articles)}            & 0.377                            \\
46                                 & ordinário \emph{(ordinary)}            & 0.313                            & 46                                 & colisão \emph{(collision)}             & 0.377                            \\
47                                 & todos   \emph{(all)}                   & 0.312                            & 47                                 & regime   \emph{(regime)}               & 0.376                            \\
48                                 & central \emph{(central)}               & 0.311                            & 48                                 & causaram \emph{(caused)}               & 0.376                            \\
49                                 & sessenta   \emph{(sixty)}              & 0.311                            & 49                                 & comarca   \emph{(county)}              & 0.374                            \\
50                                 & júri \emph{(jury)}                     & 0.310                            & 50                                 & dinheiro \emph{(money)}                & 0.373      
     \\ \hline
\end{tabular}
\end{table}

We can see that although some words are repeated in both scenarios, some words are of significant importance for a defendant's absolution or condemnation. These words can serve as a key to accurately predicting the outcome of a legal document.

One of the critical results was the difference in the number of absolutions and condemnation words: 248,460 individual word tokens for absolutions and 466,461 individual tokens for condemnations in the homicides dataset, and 66,929 individual word tokens for absolutions and 252,620 individual tokens for condemnations in the corruption dataset. Thus, judges and their clerks tend to write more in condemnation texts and reports, and thus, in both cases, try to use a more refined and unique language. This may reflect a more remarkable set of historical terms in condemnation cases, suggesting that these outcomes have specific words or that lawyers tend to write differently in this type of judicial case.

\begin{figure}[h!]
  \centering
  \includegraphics[width=\linewidth]{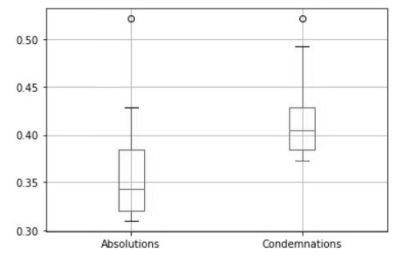}
  \caption{Boxplot of homicide word attention weights.}
  \label{fig:boxplots}
\end{figure}

Also, when we analyze the mathematical values of the weights, it can be seen that the message of condemnation weighs more heavily on the outcome of the case. Some tokens are repeated (e.g. \emph{bo} for both homicide outcomes with the same weight). However, if we look at the list, we see that the condemnation tokens continue to carry a heavy weight. In the homicides dataset, the 20th condemnation token weighs 0.42, while the 20th in the absolution list weights 0.36. In the corruption dataset, the 20th condemnation token weighs 0.96, while the 20th in the absolution list weights 0.88. From this we can see that the individual tokens on the condemnation list had a greater impact on the outcome than absolutions. This fact is consistent with previous observations and suggests that more advanced vocabulary significantly influences the text as a whole in these cases. We can see this difference graphically in Figure~\ref{fig:boxplots}. 

Looking directly at the meaning of words, we can also see patterns in the list of both results when looking at the meaning of the words. For example, in the homicide data set, in the absolution tokens, we can read many words relating to the context of the defendant, such as \emph{bem} (\emph{good}), \emph{origem} (\emph{origin}), \emph{infância} (\emph{childhoood}), \emph{social} (\emph{social}), \emph{tititi} (\emph{gossip}), \emph{cor} (\emph{color}), \emph{mãe} (\emph{mother}), \emph{anos} (\emph{years}), and \emph{soubessem} (\emph{if they knew}). Although further sociological analysis is necessary, we can conclude that many absolution texts consider the defendant's sociocultural aspects to support the decision. 

On the other hand, unique homicide condemnation tokens contain several references to violent terms, such as \emph{homicídio} (\emph{homicide}),\emph{qualificado} (\emph{aggravated}), \emph{disparos} (\emph{gun shots}), \emph{golpes} (\emph{blows}), \emph{socos} (\emph{punches}), \emph{lesões} (\emph{lesions}) and \emph{colisão} (\emph{collision}). In addition, several terms related to the judicial process, such as \emph{infração} (\emph{infraction}), \emph{penal} (\emph{criminal}), such as \emph{sentença} (\emph{sentence}), \emph{acusação} (\emph{accusation}) and \emph{comarca} (\emph{county}). In these terms, we can see that condemnation texts tend to underline the nature of crime and legal procedures to prove criminal punishment.

As for the corruption dataset, due to the shorter number of words, the analysis cannot be as precise as with the homicides dataset. However, we can see several words related to public services, such as \emph{público} (\emph{public}) and \emph{ministério} (\emph{ministry}), and related to crimes in the public sector, such as \emph{propina} (\emph{bribe}) and \emph{peculato} (\emph{embezzlement}). We can conclude that, although corruption is also present in private enterprises, in Brazil it is a crime strongly related to the public sector. 

\section{Conclusion}
\label{conc}

As an initial achievement, our work has produced a labeled corpus of judicial cases with examples of cases of homicide and corruption. Our research found that Brazilian courts did not have a labeled corpus when our study began, although some legal studies have been conducted in recent years. Since the intersection of AI and law is a discovery in Brazilian scientific research, our corpus may help future research develop new predictive strategies. Our corpus also has features we have yet to explore in this study. It may still be helpful for future analysts to find out whether factors such as the judge's gender who analyzed the case and whether the county or municipality in which the matter is located can affect the overall labeling of the document.

As another success, we demonstrate that algorithms can predict the outcome of judicial decisions from the text written in their courts in their decisions. Whether using non-neural networks such as SVM and CART or neural networks such as LSTM, GRU, and HAN, we have found results that exceed 95\% precision in most cases. Some algorithms have a high accuracy rate, but our work also proves that other methods, such as K-neighbors and RNNs, are less effective.

For models of non-neural networks, as mentioned earlier, regression trees were found to be the method that predicted results with the highest accuracy with both sets of data analyzed, often outperforming other methods considered. For example, support vector machines perform well in homicide datasets but do not match the results of regression trees in corruption data sets. In both data sets, regression trees have always maintained good predictive accuracy.

Our results match other results discovered by other research in the legal domain in many different countries, such as the one performed in \cite{Kastellec2010}, who obtained good results utilizing regression trees in the American legal system. The thesis of the author, mentioning that regression trees can study legal concepts of law and reveal patterns that other methods cannot emulate as efficiently, can also be seen in our research. The author also writes that classification trees can help increase understanding of legal rules and legal doctrine by capturing many aspects of the relationship between the facts and the outcomes of the case, a statement that could open up new opportunities for research in the field of law.

As mentioned earlier, other studies have also confirmed the effectiveness of regression trees, such as \cite{Rios-Figueroa2011}, which use the methodology to analyze the notion of judicial independence and corruption between the Supreme Courts of Latin America; \cite{Antonucci2014}, which adopted regression trees to measure the effectiveness of Italian courts, and \cite{Kufandirimbwa2012}, which used the same algorithm to predict Zimbabwe's judicial results.

As mentioned above, these studies show that while the legal system is significantly different between different languages and countries, such as Brazil, the United States, Italy, and Zimbabwe, they have similar properties that proper algorithms can accurately measure. In this way, legal texts may have intrinsic features that persist even when languages change.

Concerning the neural networks described in Section \ref{neural}, we found that GRU and HAN provide better results. The GRU provides the highest homicide dataset metrics, while hierarchical attention networks showed the highest overall accuracy overall for the corruption dataset and the second best for the homicides dataset. Our results are consistent with similar works that compare the effectiveness of HAN with other methods such as \cite{gao2018hierarchical}, who use HAN compared to Naive Bayes, logic regression, support vector machines, Random Forests, extreme gradient increases, RNN and convolutional neural networks using the technique. 

Other research has also found that HAN outperforms traditional methods. For example, \cite{ma-etal-2019-sentence} compared HAN with convolutional neural networks, SVM, LSTM, and DeClarE, and \cite{tarnpradab2018toward}, who used HAN against SVM and logistic regression models. Since these papers did not include GRU as one possible method, we cannot also conclude that this method would also be the best pick among the algorithms chosen. In addition, as mentioned previously, HAN uses an attention-based method that offers an interesting analysis of each dataset's word and sentence attention weights.

The accuracy of the GRU's results is also slightly higher than the regression tree results, suggesting that neural networks are actually effective prediction methods. However, it should be taken into account that regression trees are computationally faster than GRUs. Therefore, no one method can be selected as the best choice in each case, and every option should be tested separately. 

Our analysis of attention weights shows that words have a major impact on the meaning of the text. Although the complete sociological analysis of the word tokens is beyond the scope of our research, we can see interesting trends just by looking at the top words of each dataset, and some patterns begin to appear in the writing of the texts belonging to each of the final labels of the datasets. 

As a future research path, we intend to investigate why CART works better for judicial texts than other approaches, as shown in Section \ref{corruptiondataset}. Languages have many different characteristics, particularly those with different families, such as romantic or germanic languages. Finding that CART works better in Portuguese, Italian, and English than other methods is a remarkable discovery that deserves further research.

As another possible expansion in our scope, we plan to expand our legal text dataset, increasing the number of homicide and corruption cases and approaching other judicial topics. Adding judicial matters that are very different from those considered, such as family law or arbitrations, can significantly improve the coverage of the prediction methods developed in this investigation. We also intend to adopt other algorithms to expand the possibilities of research. Since NLP and AI are overgrowing, the speed of novelties to further improve our method is strikingly high. New algorithms and techniques can be used to get even better results.

\bibliographystyle{spbasic}      
\bibliography{main}   

\end{document}